\documentclass[10pt,twocolumn,letterpaper]{article}

\usepackage[pagenumbers]{iccv} 

\definecolor{iccvblue}{rgb}{0.21,0.49,0.74}
\usepackage[pagebackref,breaklinks,colorlinks,allcolors=iccvblue]{hyperref}

\usepackage{multirow}
\usepackage{comment}
\usepackage{bbding}

\usepackage{array}

\newlength\savewidth\newcommand\shline{\noalign{\global\savewidth\arrayrulewidth
  \global\arrayrulewidth 1pt}\hline\noalign{\global\arrayrulewidth\savewidth}}

\title{Minimizing the Pretraining Gap: Domain-aligned Text-Based Person Retrieval}

\author{Shuyu Yang$^1$ \quad 
Yaxiong Wang$^{2}\footnotemark[1]$ \quad 
Yongrui Li$^{2}$ \quad 
Li Zhu$^{1}\footnotemark[1]$ \quad 
Zhedong Zheng$^3\footnotemark[1]$ \\
$^1$Xi'an Jiaotong University \quad 
$^2$Hefei University of Technology \quad 
$^3$University of Macau\\
{\tt\small \{ysy653, wangyx15\}@stu.xjtu.edu.cn, 
hfutlyr@hfut.edu.cn,} \\
{\tt\small zhuli@mail.xjtu.edu.cn, 
zhedongzheng@um.edu.mo}
}

\begin{document}
\maketitle
\footnotetext[1]{Corresponding author.}

\begin{abstract}
In this work, we focus on text-based person retrieval, which identifies individuals based on textual descriptions. Despite advancements enabled by synthetic data for pretraining, a significant domain gap—due to variations in lighting, color, and viewpoint—limits the effectiveness of the pretrain-finetune paradigm. To overcome this issue, we propose a unified pipeline incorporating domain adaptation at both image and region levels. Our method features two key components: Domain-aware Diffusion (DaD) for image-level adaptation, which aligns image distributions between synthetic and real-world domains, \eg, CUHK-PEDES, and Multi-granularity Relation Alignment (MRA) for region-level adaptation, which aligns visual regions with descriptive sentences, thereby addressing disparities at a finer granularity. This dual-level strategy effectively bridges the domain gap, achieving state-of-the-art performance on CUHK-PEDES, ICFG-PEDES, and RSTPReid datasets. 
The dataset, model, and code are available at \url{https://github.com/Shuyu-XJTU/MRA}.
\end{abstract}


\begin{figure}[!t]
\centering
\includegraphics[width=1\linewidth]{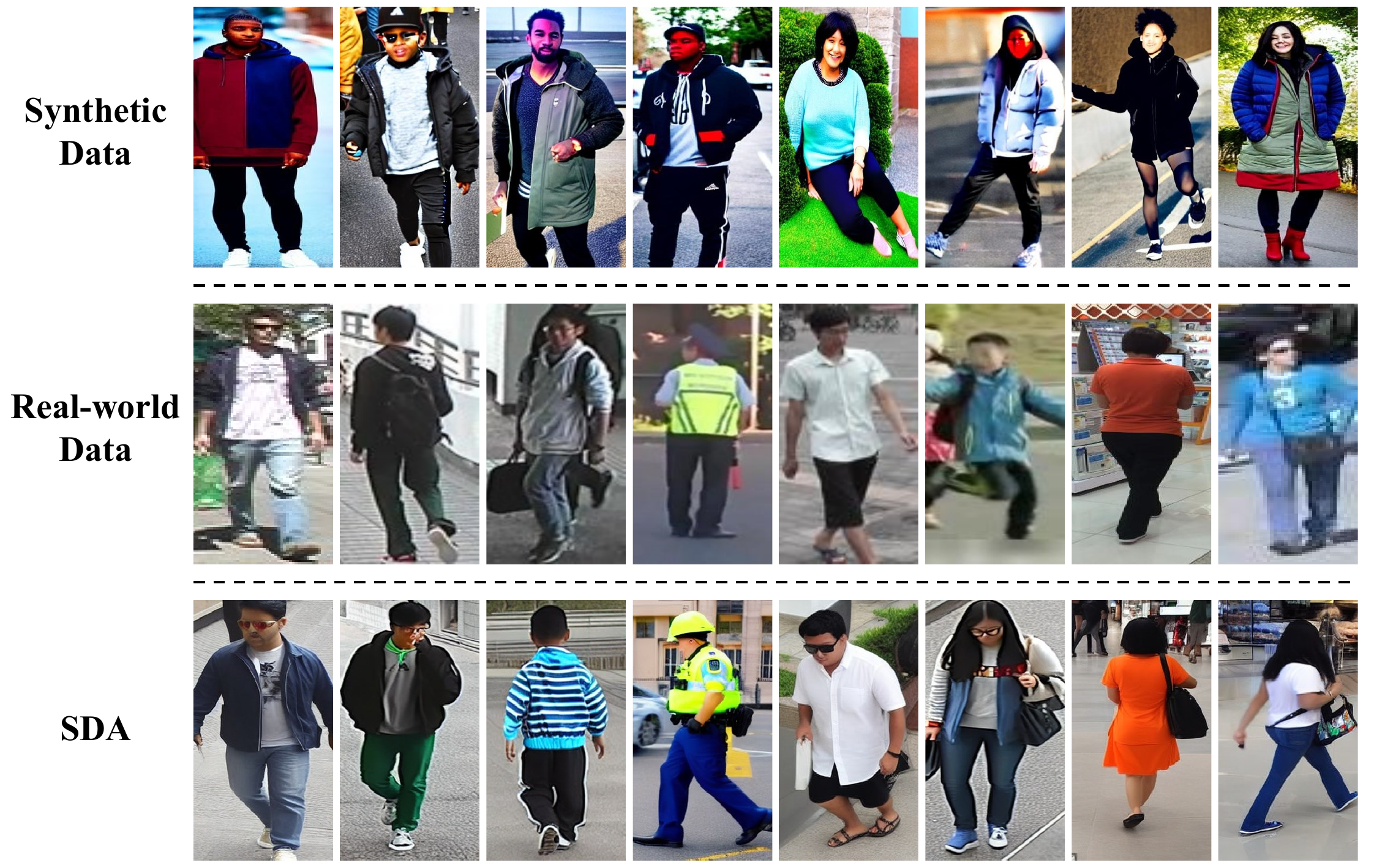}
\caption{Selected images from synthetic data generated by Diffusion model~\cite{yang2023towards}, real-world data, \ie, CUHK-PEDES~\cite{li2017person}, and our proposed Synthetic Domain-Aligned dataset (SDA).
We could observe that the visual gap between synthetic and real-world data (target domain) remains at illumination, color, viewpoints, \etc.
In contrast, images from SDA exhibit a target style while maintaining the high fidelity of the source image, characterized by a wide variety of variations in pose, appearance, background, \etc.
(Best viewed when zooming in.)
}
\label{fig: examples}
\end{figure}

\section{Introduction}
Given textual descriptions as queries, text-based person retrieval~\cite{li2017person} is to retrieve images of individuals from a large database, which is different from conventional image-based person re-identification (Re-ID)~\cite{book} and coarse-grained image-text retrieval~\cite{lei2022loopitr}.
Compared to image-based person re-identification, text-based retrieval offers a convenient and intuitive interaction for describing individual attributes due to the accessibility of textual information. 
In contrast to general cross-modal retrieval, text-based person retrieval focuses on retrieving images of individuals with finer details. 
Recently, text-based person retrieval has gained increasing attention across various domains, notably in public safety. 
Nevertheless, text-based person retrieval remains challenging due to the limited availability of data. 
As a cross-modal retrieval problem~\cite{KE2024110481, LIU2023109636}, existing pedestrian retrieval datasets typically collect images from pedestrian re-identification datasets and manually annotate the accompanying text descriptions. 
This approach significantly restricts the collection of datasets. 
On one hand, due to privacy concerns and the high costs of annotation, collecting adequate data to meet the requirements of deep learning models is difficult. 
On the other hand, manual annotation is laborious and inevitably introduces annotator bias. 
Recent studies have employed two strategies to tackle this issue. 
One line of approaches~\cite{shu2022see, jiang2023cross, bai2023rasa} involves utilizing models pretrained on extensive large-scale image-text pair datasets, thereby leveraging the robust cross-modal alignment capabilities inherent in large-scale vision language models. 
Another promising strategy~\cite{yang2023towards, zheng2017unlabeled} is to utilize generative model, \eg, GAN~\cite{creswell2018generative}, Diffusion~\cite{rombach2021highresolution, sd15}, 3D human templates~\cite{zheng2022parameter}, to generate specialized data for network pretraining. 
The strategy of generating person-text data for pretraining can well bypass the privacy issue, while generating high-quality, large-scale samples benefits from the significant advances in diffusion-based generative models.
In~\cite{yang2023towards}, Yang \etal~ make an early attempt at this thought, leveraging the off-the-shelf diffusion models to generate a large pretraining dataset, called MALS, which contains $1,510,330$ image-text pairs.
However, we observe that a significant challenge lies in the domain discrepancies that prevail between pretraining datasets and target pedestrian retrieval datasets. 
These discrepancies, stemming from varying illumination conditions, color distributions, and viewpoints, substantially undermine the efficacy of the prevailing pretraining and fine-tuning paradigm. 
For instance, as shown in Fig.~\ref{fig: examples}, there is a considerable data bias between the prevailing generated MALS~\cite{yang2023towards} and the real-world text-based pedestrian retrieval dataset CUHK-PEDES~\cite{li2017person}.

In an attempt to solve this problem, we present a unified text-based person retrieval pipeline tailored to address domain adaptation at both image and region levels. 
Our proposed pipeline comprises two integral components: Domain-aware Diffusion (DaD) and Multi-granularity Relation Alignment (MRA). 
The DaD component bridges the domain gap by diffusing the distribution of synthetically generated images from the source pretraining dataset towards the target domain, exemplified by CUHK-PEDES. 
This migration process effectively aligns the visual characteristics of the data across domains. 
In other words, Domain-aware Diffusion generates domain-aligned data for pretraining. 
The synthetic data forms the Synthetic Domain-Aligned dataset (SDA). 
We show some examples of SDA in Fig.~\ref{fig: examples}, and it is obvious that the SDA style is closer to the target domain compared to MALS. 
In Table~\ref{tab: fid}, we report the FID of MALS, GE-SD1.5, and SDA. GE-SD1.5 is a dataset consisting of synthetic images, the generation of which directly adopts Stable Diffusion 1.5~\cite{sd15}, and the prompts to guide the image generation are the text from CUHK-PEDES. The FID of SDA is about $54.85\%, 36.69\%$ lower than that of GE-SD1.5 and MALS respectively, indicating the better synthetic image quality of SDA.

To support a more fine-grained alignment during pretraining, we further augment our SDA with region-phrase level annotation. 
In specific, we detect arbitrary regions on the domain-aligned pictures based on corresponding image captions with off-the-shelf detectors~\cite{liu2023grounding}.
Based on this, we introduce  Multi-granularity Relation Alignment (MRA) to conduct a fine-grained alignment.
It encompasses the co-alignment of image-text pairs and region-phrase correspondence, thus ensuring holistic consistency across modalities. 
Going beyond image-level alignment, we delve deeper into the region level with MRA, introducing fine-grained local region matching to refine the representations further. 
This strategic design allows for better capture and alignment of intricate relations within pedestrian images and their corresponding textual descriptions. 
Extensive experiments on benchmark datasets, \ie, CUHK-PEDES, ICFG-PEDES, and RSTPReid, show that our method outperforms prior art, achieving state-of-the-art performance. 
The results underscore the effectiveness of our approach in mitigating the adverse effects of domain discrepancies and enhancing the overall retrieval accuracy. 
In summary, our primary contributions are:
\begin{itemize}
    \item We propose Domain-aware Diffusion (DaD) to bridge the domain gap at the image level by aligning the image distributions between pretraining and real-world datasets. Leveraging DaD, we construct a large-scale Synthetic Domain-Aligned dataset (SDA) comprising $1,217,750$ image-text pairs, most of which are annotated with fine-grained region-phrase information, \eg, gender, hat, and pants.
    \item Building upon SDA, we further introduce a Multi-granularity Relation Alignment (MRA) pretraining framework for region-level adaptation. MRA enhances the alignment between visual regions and their textual descriptions, thereby tackling fine-grained semantic discrepancies.
    \item Extensive experiments confirm that our dual-level adaptation approach effectively achieves domain alignment between pretraining and target datasets, surpassing the state-of-the-art methods on three real-world datasets, \ie, CUHK-PEDES, ICFG-PEDES, and RSTPReid. 
\end{itemize}

\begin{table}[!t]
\caption{The FID score of the prevailing generated dataset MALS~\cite{yang2023towards}, GE-SD1.5 by Stable Diffusion1.5, and our SDA dataset with the target dataset CUHK-PEDES. 
}
\label{tab: fid}
\small
\centering
\setlength{\tabcolsep}{7pt}
\begin{tabular}{lccc}
\hline
Dataset & GE-SD1.5~\cite{sd15} & MALS~\cite{yang2023towards}  & SDA (Ours) \\
\shline
FID $\downarrow$  & 111.41   & 92.25  & \textbf{56.56} \\
\hline
\end{tabular}
\end{table}

\begin{figure*}[!t]
\centering
\includegraphics[width=1\linewidth]{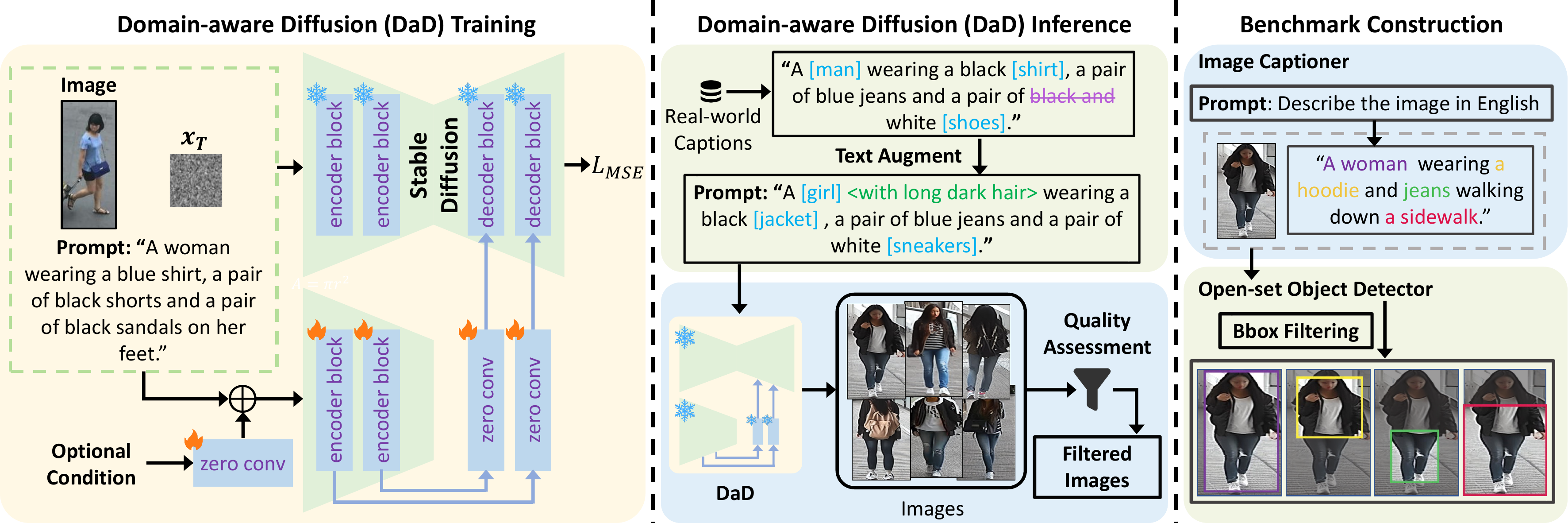}
\caption{Overview of the proposed Domain-aware Diffusion (DaD) and the Synthetic Domain-Aligned dataset (SDA) construction. 
First, we obtain DaD by fine-tuning the diffusion model on the real-world target-domain image-text pair and deploy it for accomplishing image-level domain adaptation, followed by data filtering. 
Second, we construct a synthetic pedestrian image-text pair dataset, SDA, with region annotations using off-the-shelf tools, \ie, Image Captioner and Open-set Object Detector. 
}
\label{fig: pipeline}
\end{figure*}

\section{Related Work}
\subsection{Text-based Person Retrieval}
\noindent As one of the pioneering works, Li \etal~\cite{li2017person} study the task of large-scale text-based person retrieval, introducing the seminal CUHK-PEDES benchmark. 
Echoing the broader field of image-to-text retrieval and text-to-image retrieval, early efforts focus on enhancing feature robustness~\cite{ zhang2018deep} and devising effective cross-modal alignment techniques, which include global~\cite{zheng2020dual} and local~\cite{ZHANG2025111247} alignment strategies, \eg, patch-word or region-phrase matching.
Unlike general image-text matching tasks, text-based person retrieval is enhanced by auxiliary tasks such as attribute prediction~\cite{lin2019improving} and human segmentation~\cite{wang2020vitaa}.
Specifically, three auxiliary reasoning tasks involving gender classification, appearance similarity, and image-to-text generation are developed by Zeng \etal~\cite{zeng2021relation} to facilitate alignment between pedestrian images and text. 
These approaches can be categorized into attention-based~\cite{wang2022look} and attention-free~\cite{wang2022caibc} methods.
Although attention-free techniques are generally more efficient, attention-based approaches frequently provide superior retrieval performance by effectively reducing modality gaps through cross-modal communication. 
Recent advancements in vision-language pretraining models have initiated extensive research into text-based person retrieval by refining feature discrimination~\cite{shu2022see}.
For example, utilizing the CLIP model~\cite{radford2021learning}, Yan \etal~\cite{yan2022clip} transfer the knowledge learned from large-scale generic image-text pairs. 
In~\cite{jiang2023cross} model is initialized by pretrained CLIP~\cite{radford2021learning}, while in~\cite{bai2023rasa} model is initialized by ALBEF~\cite{li2021align}. 
More recently, Yang \etal~\cite{yang2023towards} introduce a large-scale synthetic dataset named MALS for pretraining in text-based person retrieval, utilizing a joint framework of image-text matching and attribute-prompt learning. 
Subsequent fine-tuning on real-world benchmarks has significantly bolstered model performance.
Unlike the works mentioned above, we focus on the domain gap problem between the pretraining and target datasets. We propose a new text-based person retrieval pipeline, considering domain adaptation at both image and region levels.
At the image level, we tackle the domain shift between pretraining data and real-world target domain by introducing a Domain-aware Diffusion model (DaD). DaD enables the generation of a large-scale, Synthetic Domain-Aligned (SDA) dataset. In contrast to prior synthetic datasets like MALS~\cite{yang2023towards}, our SDA dataset achieves superior visual realism and is uniquely annotated with fine-grained region-phrase correspondences.
At the region level, we propose a Multi-granularity Relation Alignment (MRA) framework, which performs hierarchical alignment through the joint optimization of both global image-text pairs and local region-phrase correspondences.

\subsection{Text-to-Image Diffusion Models}
\noindent Researches into Image Diffusion Models, particularly those on Text-to-Image Diffusion, have attracted significant attention within the computer vision community. 
These advancements in Text-to-Image Diffusion Model~\cite{rombach2021highresolution} enable the creation of visually stunning images directly from text prompts, representing a significant milestone in the field. 
Latent Diffusion Models (LDM)~\cite{rombach2021highresolution} reduce computational overhead by optimizing diffusion steps in latent image space, while Stable Diffusion~\cite{sd15} provides a scalable implementation of Latent Diffusion~\cite{rombach2021highresolution}. 
Some studies attain text-guided control by adjusting prompts~\cite{hertz2022prompt}, manipulating CLIP features~\cite{kim2022diffusionclip}, and refining cross-attention mechanisms~\cite{parmar2023zero}.
Ruiz \etal ~\cite{ruiz2022dreambooth} personalize content in generated images by fine-tuning the image diffusion model with user-provided examples. 
Furthermore, ControlNet~\cite{zhang2023adding} represents a neural network architecture that enables conditional control of large pretrained Text-to-Image Diffusion models such as Stable Diffusion~\cite{sd15}, guaranteeing high-quality outputs with controls including depths, Canny edges, shape normals, Hough lines, segmentation maps, user scribbles, and human key points. 
In recent years, Diffusion models have been utilized for data augmentation~\cite{yang2024beyond}, specifically targeting coarse-grained category recognition~\cite{azizi2023synthetic} benchmarks, \eg, ImageNet~\cite{deng2009imagenet}.
However, the intricate requirements of person retrieval demand more detailed representations due to subtle individual variations. 
Consequently, in~\cite{yang2023towards}, Yang \etal ~publish the MALS dataset, which prioritizes the provision of fine-grained details crucial for text-based person retrieval.
In this paper, we introduce a Domain-aware Diffusion (DaD) component to mitigate domain differences between the datasets involved in pretraining and the target pedestrian dataset.

\begin{figure}[!t]
\centering
\includegraphics[width=1\linewidth]{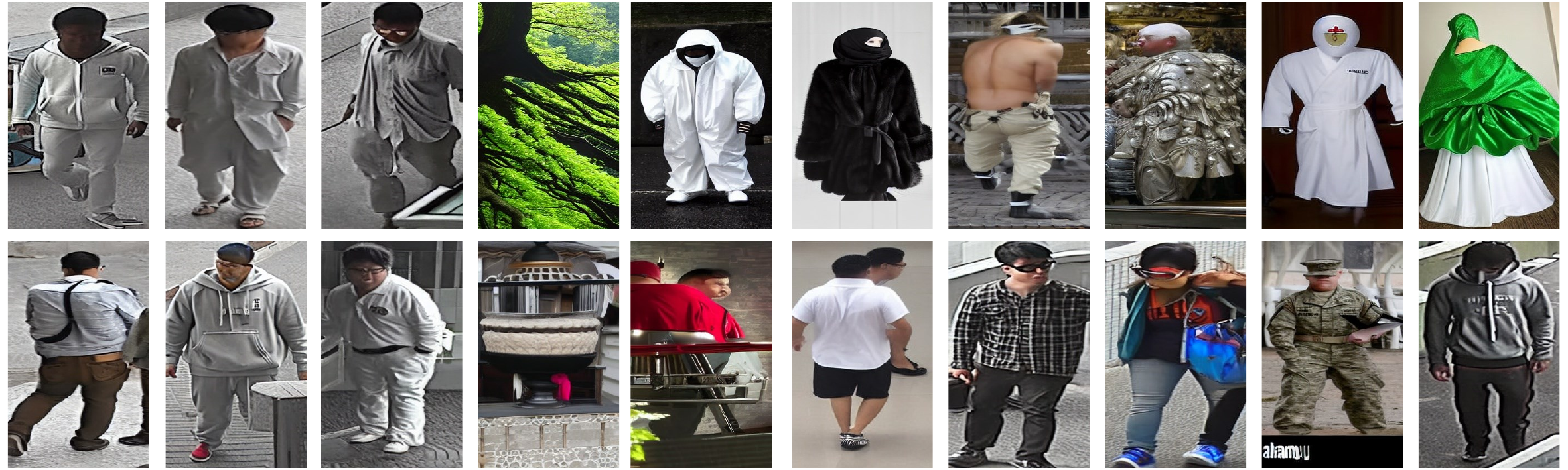}
\caption{The low-quality medium samples, generated by DaD, are mainly removed by 1) computing the file size and the mean variance of the difference between the 3 channels of every image; 2) OpenPose.
}
\label{fig: bad}
\end{figure}

\section{Benchmark} 
\label{benchmark}
\noindent As shown in Fig.~\ref{fig: pipeline}, we introduce Domain-aware Diffusion (DaD) to narrow the image-level domain gap between synthetic pretraining datasets and real-world pedestrian retrieval datasets (target domain). 
The images generated by DaD form a new dataset, the Synthetic Domain-Aligned (SDA) dataset, augmented with image captions and region-phrase annotations. 
Multi-granularity Relation Alignment (MRA) is then employed to enforce fine-grained region-level alignment during person retrieval pretraining.
In this section, we will illustrate the specific details of DaD and SDA, while the MRA workflow is elaborated in Section \ref{method}.

\subsection{Domain-aware Diffusion} 
\noindent In this paper, we use one of the prevailing real-world person retrieval datasets, \ie, CUHK-PEDES, as the target domain dataset. Previous work~\cite{yang2023towards} leverages the texts from the target domain dataset as prompts to generate images by a Diffusion Model. The synthetic images are adopted to perform pretraining. 
To migrate the domain gap between the synthetic and real-world domain, we fine-tune a Text-to-Image Diffusion Model on the target domain dataset and represent the fine-tuned model as Domain-aware Diffusion (DaD). 
In ControlNet~\cite{zhang2023adding}, the pose of the generated image can be controlled by fine-tuning Stable Diffusion~\cite{sd15}. 
Inspired by this work, we control the style of the generated image through ControlNet for image-level domain adaptation. 
Specifically, Stable Diffusion~\cite{sd15} is essentially a U-Net~\cite{ronneberger2015u} consisting of an encoder (containing 12 blocks), a middle block, and a skip-connected decoder (containing 12 blocks). 
We utilize ControlNet to create trainable copies of the 12 encoding blocks and the single middle block of the Stable Diffusion model, as illustrated in Figure~\ref{fig: pipeline}. 
During model training, the original Stable Diffusion blocks remain frozen.
The outputs of these trainable copies are added to the corresponding 12 skip-connection blocks and the middle block of the Stable Diffusion model through zero convolutions. 
The zero convolution layer is a $1 \times 1$ convolution layer with both weight and bias initialized to zeros. 
Text prompts and diffusion timesteps are encoded using the CLIP text encoder~\cite{radford2021learning} and a positional encoding-based time encoder, respectively.
We take image-text pairs from the CUHK-PEDES training set as input, set the control condition as a full white image, and optimize the model using the $\mathcal{L}_{\text{MSE}}$ (Mean Square Error) loss shown in Figure~\ref{fig: pipeline}. 
The input image resolution is set to $384 \times 256$. 
We perform fine-tuning for 3 epochs to obtain a Domain-aware Diffusion model.
There are two main advantages:
(1) Compared with traditional style transfer models, \eg, CycleGAN~\cite{zhu2017unpaired}, we deploy a fine-tuned Text-to-Image Diffusion Model for domain adaptation. 
The diffusion model shows a strong and stable ability to generate images with high content consistency to text prompts. 
The fine-tuned diffusion model can greatly reduce the gap between the generated image and the source domain data, as shown in Fig.~\ref{fig: examples}.
(2) Using synthetic images will circumvent privacy issues. By utilizing DaD as an image generator, we can produce a diverse array of pedestrian images. This approach eliminates the need to expend manpower or resources on collecting real-world pedestrian images, thereby avoiding privacy concerns and reducing labor costs.

\begin{figure*}[!t]
\centering
\includegraphics[width=0.92\linewidth]{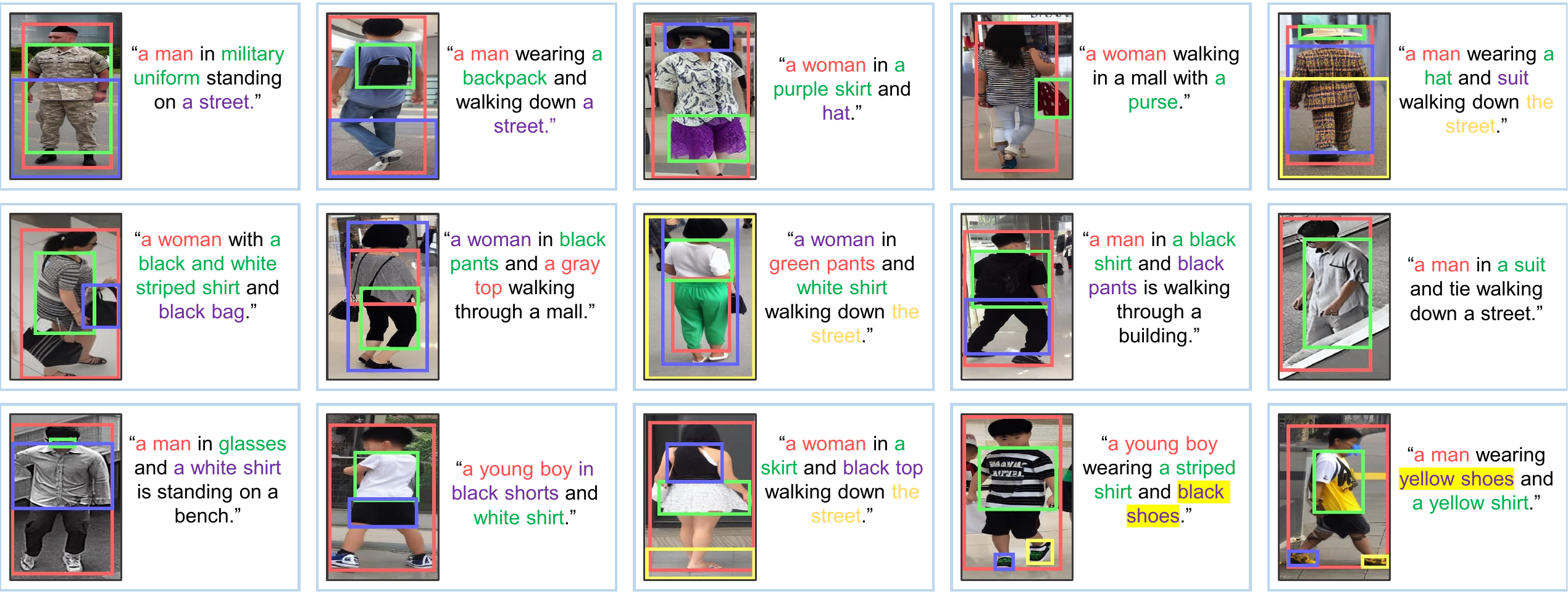}
\caption{More examples of our proposed SDA. One image-text pair usually carries 2-4 region-phrase annotations.
(Best viewed when zooming in.)}
\label{fig: sda}
\end{figure*}

\subsection{Benchmark Construction} 

\noindent\textbf{Generating Domain-adaptive Images.} 
Providing DaD with suitable textual prompts, the generative model can generate pedestrian images with a style resembling that of the target domain. 
Intuitively, we consider using the text descriptions from the CUHK-PEDES training set as a cue to guide image generation. 
To improve the image diversity, we adopt random seeds in each generation process and perform text adjustments. 
Specifically, given any text in CUHK-PEDES, we leverage EDA~\cite{wei2019eda} for text augmentation, yielding $20$ pieces of different texts (as shown in Fig.~\ref{fig: pipeline}).
EDA encompasses four augmentation techniques: synonym replacement, random insertion, random swap, and random deletion. For each CUHK-PEDES sentence, these four techniques are applied (each with $\alpha=0.5$) to generate five augmented sentences, separately. The resulting 20 texts are then randomly shuffled, and the first 19 augmented sentences are selected. Finally, the original sentence is added to obtain a total of 20 distinct texts.
As a result, $68,126$ CUHK-PEDES texts are converted into 20-fold CUHK-PEDES prompts.
The adjusted CUHK-PEDES prompts are input to generate images. 
Following the above steps, we obtain new domain-adaptive images.

\noindent\textbf{Image Filtering.} 
To further improve the quality of the generated data, we design some data filtering strategies, as shown in Fig.~\ref{fig: pipeline}.
First, grayscale images are filtered by calculating the mean-variance of the difference between the 3 channels of each picture. 
Next, noise images, which contain distorted portraits, are filtered by OpenPose~\cite{cao2017realtime}.
In Fig.~\ref{fig: bad}, we show some low-quality picture samples.
Although $10.63\%$ of low-quality images are filtered, oversaturation still exists in some of the filtered images due to limitations of the diffusion model itself.

\noindent\textbf{Caption Diversifying.} 
Obtaining high-quality domain-aligned images, we attempt to use the pairs of synthetic images and CUHK-PEDES prompts directly for pretraining. The results are not satisfactory. This limitation arises because every twenty prompts are derived from the same text description, resulting in a lack of textual diversity.
Our solution is to adopt the pretrained Image Captioner model BLIP2 ~\cite{li2023blip} to re-generate the corresponding text descriptions for each synthetic image, yielding new image-text pairs, as shown in Fig.~\ref{fig: pipeline}.
It is worth noting that while BLIP2 provides us with diverse texts, this process may introduce inherent biases present in the BLIP2 model itself.

\noindent\textbf{Region-Phrase Annotation Generation.} 
Many text-based person retrieval works~\cite{shu2022see, yang2023towards, jiang2023cross} have validated that finer-grained features tend to help better image-text alignment. 
Inspired by these works, we further add region annotations to involve multi-granularity visual concepts and corresponding texts. 
Considering the high cost of manual annotation, we employ the Open-set Object Detector Grounding DINO~\cite{liu2023grounding}, which can detect arbitrary objects through human inputs such as category names or reference expressions. 
Specifically, we input image-text pairs into the trained Grounding DINO to obtain region annotations, including bounding boxes, bounding box confidence, and phrase descriptions of the bounding box region.
When Grounding DINO performs the detection, we set both the text threshold and the box threshold to $0.35$.
For each image-text pair, more than two region annotations can be obtained in most cases.

\begin{figure*}[!t]
\centering
\includegraphics[width=1\linewidth]{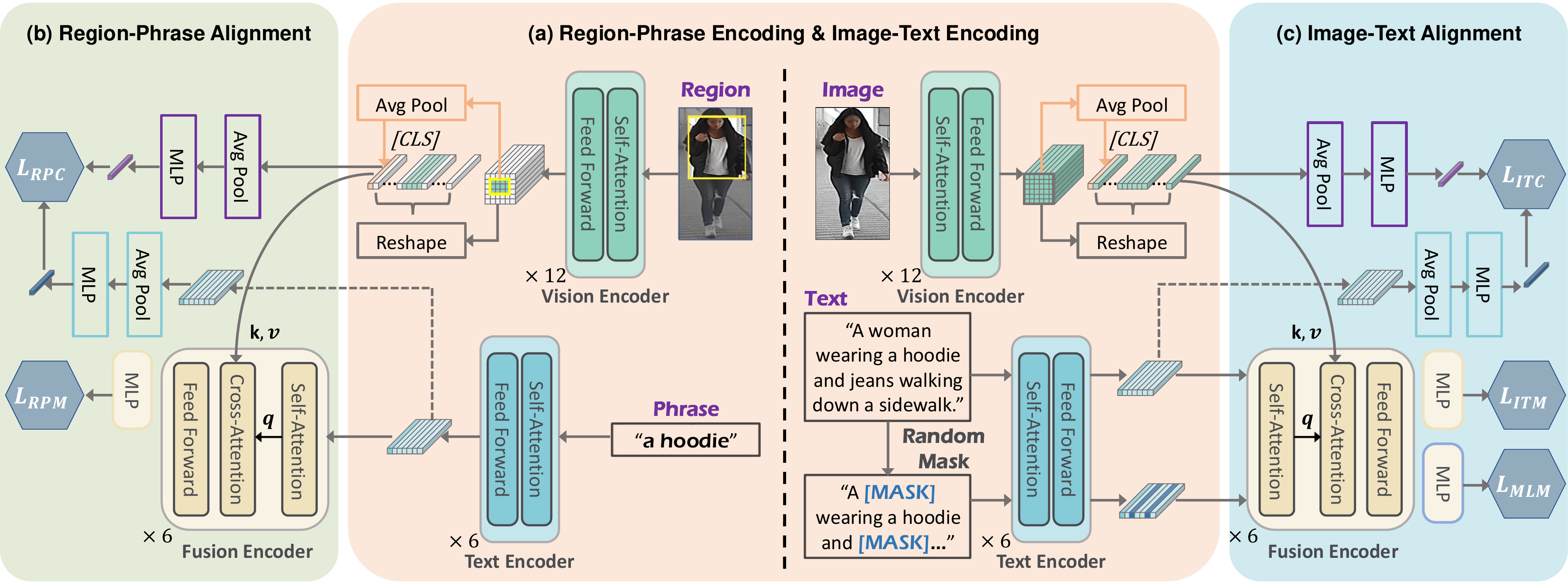}
\caption{Overview of the proposed Multi-granularity Relation Alignment framework (MRA). MRA first conducts (a) region-phrase encoding and image-text encoding by the shared Vision Encoder ($E_V$) and the shared Text Encoder ($E_T$). The model is constrained with cross-modal alignment at both the (b) region-phrase and (c) image-text levels, where the shared Fusion Encoder ($E_F$) seeks to fuse the vision and text embeddings for the subsequent predictions. 
}
\label{fig: framework}
\end{figure*}

\subsection{Synthetic Domain-Aligned Benchmark} 
\noindent Following the above steps, a large-scale diverse benchmark, \ie, Synthetic Domain-Aligned dataset (SDA), for text-based person retrieval pretraining is built.
In Fig.~\ref{fig: sda}, we show some image-text pairs annotated with region-phrase pairs from SDA. 
The style of SDA images is similar to CUHK-PEDES.
SDA contains $1,217,750$ image-text pairs, roughly $18 \times$ more than the $68,126$ in the CUHK-PEDES training set.
To the best of our knowledge, this is the first pedestrian dataset with region annotations. 
Most of the existing datasets have attribute annotations but provide no information about the location of the relevant attributes in the image.
In SDA, different-grained visual concepts and language description pairs refer to coarse-grained image-text pairs and fine-grained image region-phrase pairs. 
In other terms, an image can contain several different visual concepts, each of which has a corresponding matching textual description, denoted as $(I, T, \{(I_R^i, T_R^i)\}^M_{i=1})$.
$(I, T)$ represents an image-text pair, while $(I_R^i, T_R^i)$ denotes a region-phrase pair.
$I_R^i$ is an image region, which is defined by the bounding box $b^i=(cx, cy, w, h)$. $cx, cy, w, h$ denote the center coordinates, width, and height of the box, respectively. 
When a complete picture represents a visual concept by itself, its bounding box is $(0.5, 0.5, 1, 1)$.
$T_R^i$ is the corresponding textual annotations, \ie, a phrase describing the content of the region $I_R^i$. 
Note that some image-text pairs may not produce region annotations, \ie, $M = 0$.

\noindent\textbf{Discussion: The contribution to the community.} 
The proposed benchmark follows a similar synthetic manner as existing synthetic pretraining datasets~\cite{yang2023towards,zheng2017unlabeled}, which mitigates the privacy concerns and partially solves the data scarcity for the data-hungry models. 
However, our contribution diverges from these prior works in two significant dimensions: 
\textbf{1) Real-world Image Style.} The visual characteristics of our dataset are meticulously designed to emulate those found in actual pedestrian datasets, thereby promoting effective transfer learning from the pretraining phase to fine-tuning (on real-world data). 
As shown in Section~\ref{experiment}, our approach surpasses APTM~\cite{yang2023towards}, even though it leverages MALS, a more voluminous corpus encompassing $1.51$ million image-text pairs (approximately $0.3$ million more than SDA). 
\textbf{2) Fine-Grained Region-Phrase.} We introduce detailed region-phrase annotations, a critical factor for advancing cross-modal retrieval capabilities and refining the model aptitude for distinguishing subtle visual cues. 
To the best of our knowledge, SDA is one of the pioneering datasets to provide such annotations, whether within the scope of Re-ID or text-based pedestrian retrieval tasks. 
Although some datasets offer attribute information, they often neglect to specify the spatial location of these attributes within the images. 
By offering these precise annotations, SDA empowers researchers to innovate, particularly in improving the model proficiency in recognizing intricate details, thus propelling advancements in fine-grained person retrieval.

\section{Methodology}
\label{method}
\noindent Building upon the multi-granular annotations in SDA, we propose the Multi-granularity Relation Alignment (MRA) framework. 
As depicted in Fig.~\ref{fig: framework}, MRA is optimized through two complementary constraints:
1) fine-grained alignment between image regions and description phrases via Region-Phrase Contrastive (RPC) Learning and Region-Phrase Matching (RPM) Learning; 
2) coarse-grained alignment between full images and texts through Image-Text Alignment.

\subsection{Region-Phrase Contrastive Learning}
\noindent Region-Phrase Contrastive (RPC) Learning aligns the region-phrase pair $(I_R, T_R)$ with contrastive learning.
Fig.~\ref{fig: framework} shows that the $N$ image-text pairs and the $N$ corresponding region-phrase pairs are fed into MRA in a training batch $D$.
For $N$ pairs of region and phrase $\{(I_R^1, T_R^1), (I_R^2, T_R^2), ...\}^N$, $(I_R^j, T_R^j)$ is a positive match, while $(I_R^j, T_R^{\bar{j}})$ is a negative match ($j, \bar{j} \in [1, N], j \neq \bar{j}$). 
We first obtain the region embeddings $E_V(I_R)$ and phrase embeddings $E_T(T_R)$ through the Vision Encoder $E_V$ and the Text Encoder $E_T$.

\noindent\textbf{Region Encoding.}
Following the existing work~\cite{yang2023towards}, the Swin Transformer (Swin-Base)~\cite{liu2021swin} is deployed as the Vision Encoder $E_V$. Feeding the region to $E_V$ is a straightforward strategy for region encoding; however, this fashion is not friendly to the small-sized region. In addition, the sole region lacks the surrounding contexts for better representation. With this consideration, we resort to indexing the region features from the full image features. 
Formally, given the region $I_R$, its corresponding image $I$ is evenly divided into $N^I$ non-overlapping patches (with a patch size of $32 \times 32$) of fixed size first. 
The image size is set to $224 \times 224$, and $N^I = 49$.
Then, the sequence of $N^I$ patches is fed into the Transformer block layers of $E_V$, outputting $N^I$ high-dimensional embeddings.
Region embedding $E_V(I_R)$ derives from $V$: $E_V(I_R)= \{v^{i} | \text{ if } v^{i} \text{ overlaps with } I_R, i=1,2,\dots,N^I\}$.
Then the mean feature of $E_V(I_R)$ is calculated and used as the \texttt{[CLS]} patch embedding of $I_R$, which is together with the patch features in $E_V(I_R)$ to form the full representation of the region $I_R$.

\noindent\textbf{Phrase Encoding.}
Without loss of generality, we use the first 6 layers of BERT as the Text encoder $E_T$ to extract the textual representation. 
Specifically, for a given phrase $T_R$, it is first tokenized by a BERT Tokenizer with a vocabulary size of $30,522$. 
A \texttt{[CLS]} token is added at the beginning of the set of tokens.
The tokens are then passed into the six Transformer layers of BERT to obtain the textual embedding $E_T(T_R)$.

The obtained region and phrase embeddings are first averaged separately and mapped to the low-dimensional region features by different MLPs. 
Then, we predict the similarity between regions and phrases by calculating the cosine similarity $s(\cdot,\cdot)$ between region and phrase. 
Given a pair of region and phrase $(I_R, T_R)$, the region-to-phrase similarity is defined as follows:
\begin{equation}
\label{eq: S_R2P}
S_{\text{R2P}} = \frac{\exp(s(I_R, T_R)/\tau)}{\sum_{i=1}^{N}\exp(s(I_R, T_R^i)/\tau)},
\end{equation}
where we still use $I_R, T_R$ to represent the respective features for brevity, and $\tau$ is a learnable temperature parameter.
Similarly, the phrase-to-region similarity is $S_{\text{P2R}}$.
Finally, the RPC loss function is defined as:
\begin{equation}
\label{eq: L_PRC}
\begin{split}
\mathcal{L}_{\text{RPC}} = -\frac{1}{2} \mathbb{E}[\log S_{\text{R2P}} + \log S_{\text{P2R}}].
\end{split}
\end{equation}

\subsection{Region-Phrase Matching Learning.}
\noindent Region-Phrase Matching (RPM) Learning is a binary classification task used to further differentiate between positive (matched) and negative (unmatched) region-phrase pairs. 
If a negative sample is randomly selected to form a negative sample pair, the classification problem would be too easy for the model. 
Therefore, based on RPC, we form negative sample pairs by selecting one hard-negative sample for each region and phrase through a Hard-Negative Sampling strategy. 
Specifically, for a given region $I_R$, a Hard-Negative Sample $\Bar{T}_R$ is randomly selected in proportion to the similarity score in Eq. \ref{eq: S_R2P}. 
Similarly, a hard-negative sample for each phrase can be obtained analogously. 
Then, the $N$ matched region-phrase pairs and $2N$ corresponding mismatched region-phrase pairs are fed into the Fusion Encoder $E_F$ together to obtain the fusion embeddings, as shown in Fig. \ref{fig: framework}.

\noindent\textbf{Fusion Encoder $E_F$.}
We adopt the last six layers of BERT as $E_F$. 
$E_F$ takes phrase embedding as inputs and fuses region embedding at each cross-attention module of the Transformer Blocks. 
Specifically, the phrase embedding is used as a query (q in Fig. \ref{fig: framework}), and the region embedding works as key (k) and value (v). 
The fusion embedding yielded by the Fusion Encoder is used to perform further cross-modal alignment tasks.
We project the \texttt{[CLS]} embeddings of the fused embeddings into the two-dimensional space by MLP to get the predicted matching likelihood $\hat{p}_m(I_R, T_R)$.

The Region-Phrase Matching (RPM) loss is obtained by calculating the cross-entropy loss between the predicted matching likelihood and the ground-truth matching likelihood $p_m(I_R, T_R)$:
\begin{equation}
\label{eq: L_PRM}
\begin{split}
\mathcal{L}_{\text{RPM}} =& - \mathbb{E}[p_{m}(I_R, T_R)\log \hat{p}_{m}(I_R, T_R)\\
&+ (1-p_{m}(I_R, T_R))\log (1-\hat{p}_{m}(I_R, T_R))],
\end{split}
\end{equation}
\noindent where $p_{m}$ is a ground-truth binary label. If the image and text are true-matches, $p_{m} = 1$, otherwise $p_{m} = 0$.

\noindent\textbf{Discussion. Why do we need region-phrase level alignment?}
While recent methods for image-text matching have productively incorporated both global and local feature representations \cite{jiang2023cross, bai2023rasa, yang2023towards, yan2022clip}, a systematic, explicit alignment between discrete image regions and their corresponding textual phrases remains largely unexplored. This constitutes a significant limitation, as fine-grained cross-modal understanding fundamentally requires the precise grounding of visual constituents within their descriptive linguistic contexts. Our work directly bridges this gap by introducing an explicit region-phrase alignment mechanism, facilitated by the novel structured annotations in our proposed SDA dataset.
As shown in Table~\ref{tab: abl}, Method M1, which relies solely on global image-text alignment, exhibits a marked performance shortfall relative to our complete MRA framework. This comparative analysis confirms that integrating region-phrase alignment contributes a substantial and quantifiable improvement to retrieval accuracy.
Furthermore, our investigation into even finer-grained object-word alignment (variants 1 and 2 in Table~\ref{tab: abl}) yields a critical observation: exceeding an optimal level of granularity can be counterproductive. The suboptimal performance of object-word alignment suggests that excessive fragmentation of the representation may introduce noise and disrupt the compositional semantics preserved by phrase-level descriptions. This evidence highlights that region-phrase alignment achieves an effective balance, providing the necessary discriminative detail to surpass global methods while avoiding the loss of contextual coherence inherent in over-segmented approaches.

\begin{figure*}[!t]
\centering
\begin{minipage}[t]{0.45\linewidth}
    \centering
    \small    \makeatletter\def\@captype{table}\makeatother
    \caption{Performance comparisons with state-of-the-art methods on the CUHK-PEDES dataset.}
    \label{tab:sota_CUHK}
    \centering
    \resizebox{1\linewidth}{!}{ 
        \renewcommand\arraystretch{1} 
        \begin{tabular}[h!]{p{2.6cm}|m{1.2cm}<{\centering}m{1.2cm}<{\centering}m{1.2cm}<{\centering}m{1.2cm}<{\centering}}
            \shline
            \textbf{Method} & \textbf{R@1} & \textbf{R@5} & \textbf{R@10} & \textbf{mAP}  \\
            \hline
            CNN-RNN~\cite{reed2016learning} & 8.07 & - & 32.47 & - \\ 
            GNA-RNN~\cite{li2017person} & 19.05 & - & 53.64 & - \\ 
            PWM-ATH~\cite{chen2018} & 27.14 & 49.45 & 61.02 & - \\ 
            GLA~\cite{chen2018improving} & 43.58 & 66.93 & 76.2 & - \\
            Dual Path~\cite{zheng2020dual} & 44.40 & 66.26 & 75.07 & - \\
            CMPM+CMPC~\cite{zhang2018deep} & 49.37 & - & 79.21 & - \\
            MIA~\cite{niu2020improving} & 53.10 & 75.00 & 82.90 & -\\
            A-GANet~\cite{liu2019deep} & 53.14 & 74.03 & 81.95 & - \\  
            ViTAA~\cite{wang2020vitaa} & 55.97 & 75.84 & 83.52 & 51.60 \\ 
            IMG-Net~\cite{wang2020img} & 56.48 & 76.89 & 85.01 & - \\
            CMAAM~\cite{aggarwal2020text} & 56.68 & 77.18 &	84.86 & - \\
            HGAN~\cite{zheng2020hierarchical} &	59.00 & 79.49 & 86.62 & - \\
            DSSL~\cite{zhu2021dssl} & 59.98 & 80.41 & 87.56 & - \\
            MGEL~\cite{wang2021text} & 60.27 & 80.01 & 86.74 & - \\
            SSAN~\cite{ding2021semantically} & 61.37 & 80.15 & 86.73 & - \\
            TBPS~\cite{han2021text} & 61.65 &80.98 & 86.78 & - \\
            TIPCB~\cite{chen2022tipcb} & 63.63 &82.82 & 89.01 & - \\
            LBUL~\cite{wang2022look} & 64.04 &82.66 & 87.22 & - \\
            CAIBC~\cite{wang2022caibc} & 64.43 & 82.87 &88.37 & - \\
            AXM-Net~\cite{farooq2022axm} & 64.44 & 80.52 & 86.77 & 58.73 \\
            SRCF~\cite{suo2022simple} & 64.88 &83.02 & 88.56 & - \\
            LGUR~\cite{shao2022learning} & 65.25 & 83.12 & 89.00 & - \\
            CFine~\cite{yan2022clip} & 69.57 & 85.93 & 91.15 & - \\
            IRRA~\cite{jiang2023cross} & 73.38 & 89.93 & 93.71 & 66.13 \\
            SAMC~\cite{lu2024mind} & 74.03 & 89.18 & 93.31 & 68.42 \\
            RaSa~\cite{bai2023rasa} & 76.51 & 90.29 & 94.25 & \textbf{69.38} \\
            APTM~\cite{yang2023towards} & 76.53	& 90.04 & 94.15 & 66.91\\
            \hline
            Baseline & 71.23 & 87.54 & 92.43 & 64.75 \\
            Ours & \textbf{77.21}	& \textbf{90.66} & \textbf{94.46} & 68.50 \\
            \shline
        \end{tabular}
    }    
\end{minipage}
\hspace{0.4cm}
\quad
\begin{minipage}[t]{0.45\linewidth}
    \centering
    \begin{minipage}[t]{1.0\linewidth}
        \centering
        \small
        \makeatletter\def\@captype{table}\makeatother
        \caption{Performance comparisons with state-of-the-art methods on the ICFG-PEDES dataset.} 
        \label{tab:sota_ICFG}
        \centering
        \resizebox{1\linewidth}{!}{
            \renewcommand\arraystretch{1} 
            \begin{tabular}[h!]{p{3cm}|m{1.2cm}<{\centering}m{1.2cm}<{\centering}m{1.2cm}<{\centering}m{1.2cm}<{\centering}} 
                \shline
                \textbf{Method} & \textbf{R@1} & \textbf{R@5} & \textbf{R@10} & \textbf{mAP}  \\
                \hline
                Dual Path~\cite{zheng2020dual}   & 38.99 & 59.44 & 68.41 & - \\
                CMPM+CMPC~\cite{zhang2018deep}   & 43.51 & 65.44 & 74.26 & - \\
                MIA~\cite{niu2020improving}      & 46.49 & 67.14 & 75.18 & - \\
                SCAN~\cite{lee2018stacked}       & 50.05 & 69.65 & 77.21 & - \\
                ViTAA~\cite{wang2020vitaa}       & 50.98 & 68.79 & 75.78 & - \\
                SSAN~\cite{ding2021semantically} & 54.23 & 72.63 & 79.53 & - \\
                IVT~\cite{shu2022see}            & 56.04 & 73.60 & 80.22 & - \\
                LGUR~\cite{shao2022learning}     & 59.02 & 75.32 & 81.56 & - \\
                CFine~\cite{yan2022clip}         & 60.83 & 76.55 & 82.42 & - \\
                IRRA~\cite{jiang2023cross}       & 63.46 & 80.25 & 85.82 & 38.06 \\
                SAMC~\cite{lu2024mind} & 63.68 & 79.69 & 85.21 & \textbf{42.41} \\
                RaSa~\cite{bai2023rasa}          & 65.28 & 80.40 & 85.12 & 41.29 \\
                APTM~\cite{yang2023towards}      & 68.51 & 82.99 & 87.56 & 41.22\\
                \hline
                Baseline & 66.02 & 81.67 & 86.78 & 39.04 \\ 
                Ours & \textbf{68.93} & \textbf{83.46} & \textbf{88.07} & 41.38 \\
                \shline
            \end{tabular}
        }
    \end{minipage}    
    \begin{minipage}[t]{1.0\linewidth}
        \centering
        \small
        \makeatletter\def\@captype{table}\makeatother
        \caption{Performance comparisons with state-of-the-art methods on the RSTPReid dataset.}
        \label{tab:sota_RSTP} 
        \centering
        \resizebox{1\linewidth}{!}{
            \renewcommand\arraystretch{1} 
            \begin{tabular}[h!]{p{3cm}|m{1.2cm}<{\centering}m{1.2cm}<{\centering}m{1.2cm}<{\centering}m{1.2cm}<{\centering}}
                \shline
                \textbf{Method} & \textbf{R@1} & \textbf{R@5} & \textbf{R@10} & \textbf{mAP}  \\
                \hline
                DSSL~\cite{zhu2021dssl}     & 32.43 & 55.08 & 63.19 & - \\
                LBUL~\cite{wang2022look}    & 45.55 & 68.20 & 77.85 & - \\
                IVT~\cite{shu2022see}       & 46.70 & 70.00 & 78.80 & - \\
                CAIBC~\cite{wang2022caibc}  & 47.35 & 69.55 & 79.00 & - \\
                CFine~\cite{yan2022clip}    & 50.55 & 72.50 & 81.60 & - \\
                IRRA~\cite{jiang2023cross}  & 60.20 & 81.30 & 88.20 & 47.17 \\
                SAMC~\cite{lu2024mind} & 60.80 & 82.35 & 89.00 & 49.67 \\
                RaSa~\cite{bai2023rasa}     & 66.90	& \textbf{86.50} & 91.35 & 52.31 \\
                APTM~\cite{yang2023towards} & 67.50	& 85.70 & \textbf{91.45} & 52.56 \\
                \hline
                Baseline & 56.30 & 77.55 & 85.20 & 45.88 \\
                Ours & \textbf{68.15} & 86.30 & 91.10 & \textbf{53.77}\\
                \shline
            \end{tabular}
        }
    \end{minipage}  
\end{minipage}  
\end{figure*}

\subsection{Image-Text Alignment}
\noindent In addition to region-phrase level alignment (RPC and RPM), we combine the tasks of Image-Text Contrastive (ITC) learning, Image-Text Matching (ITM) learning, and Masked Language Modeling (MLM) to constrain the model to align pedestrian images and text descriptions.

\noindent\textbf{Image-Text Contrastive (ITC) Learning} is performed in a similar fashion of RPC. The ITC loss $\mathcal{L}_{\text{ITC}}$ is obtained following Eq.\ref{eq: S_R2P}-\ref{eq: L_PRC}.

\noindent\textbf{Image-Text Matching (ITM) Learning.}
The goal of Image-Text Matching (ITM) Learning is to predict whether the image-text pairs are positive pairs or not.
The ITM loss $\mathcal{L}_{\text{ITM}}$ is computed similarly using Eq. \ref{eq: L_PRM}.

\noindent\textbf{Masked Language Modeling (MLM).}
Masked Language Modeling (MLM) predicts masked words using information from matched image-text pairs as clues. 
In this paper, we randomly mask out text tokens with a probability of $25\%$. 
Among the masked tokens: 
1) $10\%$ are replaced by random text tokens; 
2) $80\%$ are replaced by special text tokens \texttt{[MASK]}; 
3) the remaining $10\%$ stay unchanged.
Given a matched $(I, T)$, mask $T$ as $\hat{T}$ and use $\hat{f}^i$ to denotes fused embedding of masked token $\hat{t}^i$.
We predict what the token $\hat{t}^i$ should be by a classified MLP header, denoted as $\hat{p}_{\text{mask}}^i(I, \hat{T}) = \text{Softmax}(\text{MLP}(\hat{f}^i))$.
$p_{\text{mask}}^i$ is a one-hot vector representing the ground-truth probability. Then the MLM loss can be formulated as follows by the cross-entropy:
\begin{equation}
\label{eq: L_MLM}
\mathcal{L}_{\text{MLM}} = -\mathbb{E}[p_{\text{mask}}^i(I, \hat{T}) log( \hat{p}_{\text{mask}}^i(I, \hat{T}))].
\end{equation}
MRA jointly optimizes Region-Phrase Alignment (RPC and RPM) and Image-Text Alignment. 
The overall MRA loss is:
\begin{equation}
\label{eq: L_MRA}
\mathcal{L}_{\text{MRA}} = \mathcal{L}_{\text{ITC}} + \mathcal{L}_{\text{ITM}} +\mathcal{L}_{\text{MLM}} + \beta (\mathcal{L}_{\text{RPC}} + \mathcal{L}_{\text{RPM}}),
\end{equation}
\noindent where $\beta$ denotes the weight of RPC and RPM loss.

\section{Experiment}
\label{experiment}
\subsection{Experimental Setup}

\begin{figure*}[!t]
\centering
\includegraphics[width=0.72\linewidth]{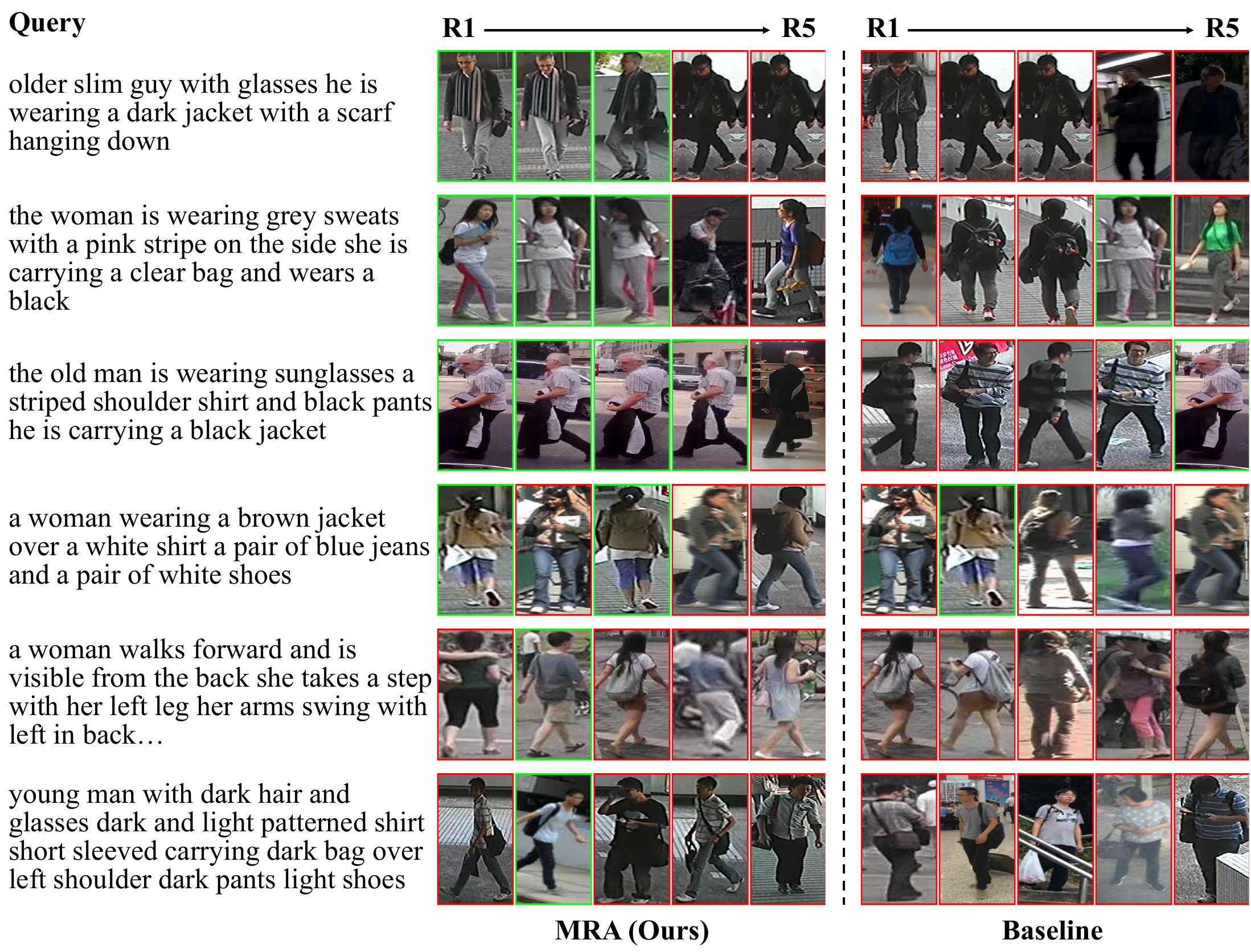}
\caption{Qualitative text-based person retrieval results on CUHK-PEDES. The results are ranked from left to right in descending order of matching score. Green and red bounding boxes indicate correct and incorrect matches, respectively. 
The first four lines of retrieval results are success cases, while the bottom two lines are failures.
The proposed MRA method successfully recalls more true positives at the top rankings compared to the baseline. Notably, the failure cases of MRA are semantically reasonable, as the incorrect retrievals still correspond meaningfully to the text query.
}
\label{fig: result}
\end{figure*}

\noindent\textbf{Datasets.} 
\noindent We evaluate our approach on three public text-based person retrieval datasets, \ie, \textbf{CUHK-PEDES}~\cite{li2017person}, \textbf{ICFG-PEDES}~\cite{ding2021semantically}, and \textbf{RSTPReid}~\cite{zhu2021dssl}. 
In particular, \textbf{CUHK-PEDES}~\cite{li2017person} includes $80,440$ description sentences and $40,206$ photos of $13,003$ people. It is divided into three subsets: a training set that contains $34,054$ images and $68,126$ description sentences; a validation set that contains $3,078$ images and $6,158$ description sentences; and a testing set that contains $3,074$ images and $6,156$ description sentences. 
\textbf{ICFG-PEDES}~\cite{ding2021semantically} 
is incubated from MSMT17~\cite{wei2018person} and 
has $54,522$ images of $4,102$ individuals. There is an average of $37.2$ words in each description sentence for each image. The training subset has $34,674$ image-text pairs for $3,102$ people, whereas the testing subset has $19,848$ image-text pairs for the remaining $1,000$ people.
\textbf{RSTPReid}~\cite{zhu2021dssl} comprises $20,505$ images of $4,101$ people
and is also created by compiling MSMT17~\cite{wei2018person} data. 
Every image is associated with two sentences, each of which contains at least $23$ words. More specifically, there are $3,701$ IDs in the training set while $200$ and $200$ IDs in the validation and testing sets, respectively.

\noindent\textbf{Evaluation Metrics.} 
For the text-based person retrieval, following earlier studies, we report recall rates, \eg, R@1, R@5,  R@10, and mAP.  The mean average precision (mAP) is the average area under the precision-recall curve across all queries. The higher Recall rate and mAP indicate better retrieval performance.

\noindent\textbf{Implementation Details.}
During pretraining, we optimize MRA with Pytorch on $4$ NVIDIA GeForce RTX 3090 GPUs for 32 epochs, and the mini-batch size is $40$. 
The learning rate is set to $5e^{-5}$ and is decayed to $5e^{-6}$ following a linear schedule, while the first $2600$ training iterations are performed as a warm-up. 
We use the AdamW~\cite{loshchilov2018decoupled} optimizer with a weight decay of $0.01$.
Before the training starts, the Vision Encoder, Text Encoder, and Fusion Encoder are initialized with Swin Transformer$_{\text{base}}$~\cite{liu2021swin}, the first six and the last six layers of BERT$_{base}$~\cite{devlin2019bert}, respectively. 
The trainable parameters of MRA are $226.5M$.
MRA adopts the image resolution of $224 \times 224$, while Random Horizontal Flipping and random Erasing are used to augment the image. 
The maximum text token length is set to $56$. 
Pretraining takes about 4 days.

\noindent\textbf{Fine-tuning Details.}
We fine-tune the model on the pedestrian retrieval datasets and only optimize ITC, ITM, and MLM.
During fine-tuning, we adopt $4$ NVIDIA A100 40GB GPUs to train the model for $30$ epochs with a mini-batch size of $40$. 
The learning rate following a linear schedule is equal to $1e^{-4}$ on CUHK-PEDES and ICFG-PEDES, $5e^{-5}$ on RSTPReid. 
The image is resized to $384 \times 256$ and we deploy Random Horizontal Flipping and Random Erasing for image data augmentation, while EDA~\cite{wei2019eda} for text data augmentation.

\begin{table*}[!t]
\caption{Performance Comparison on image-based person re-identification dataset Market-1501~\cite{zheng2015scalable}.
}
\label{tab: reid}
\small
\centering
\begin{tabular}{l|c|c|cccc}
\shline
Method & Pretraining Dataset & \# Data & R@1 & R@5 & R@10 & mAP \\
\hline
Swin Transformer   & ImageNet-1K~\cite{deng2009imagenet} & 1.28M & 91.12 & 96.62 & 98.22 & 76.31 \\
Swin Transformer & SDA & 1.22M & \textbf{94.39} & \textbf{97.86} & \textbf{98.63} & \textbf{84.26} \\
\shline
\end{tabular}
\end{table*}

\subsection{Comparison with Competitive Methods}
\noindent \textbf{Quantitative Results.} 
In this section, we present the results of the comparison with state-of-the-art methods on three person retrieval datasets. 
In other words, we adapt MRA for pedestrian retrieval and evaluate the effectiveness of our method on CUHK-PEDES, ICFG-PEDES, and RSTPReid datasets.
In reference, we first obtain the embeddings of all images and texts in the test set, followed by computing the image-to-text similarity. 
For each text, the top $512$ image candidates with the highest similarity to it are selected. 
Then the candidates are re-ranked by the Fusion Encoder. 
As shown in Table~\ref{tab:sota_CUHK}, \ref{tab:sota_ICFG}, \ref{tab:sota_RSTP}, our proposed method achieves SOTA recall rate on all three datasets. 
Note that the Baseline model denotes the model obtained without pretraining and training on the three datasets directly (only ITC, ITM, and MLM is optimized).
On CUHK-PEDES, the R@1, R@5, and R@10 of MRA are $77.21 \%$, $90.66 \%$, and $94.46 \%$ respectively, outperforming the second-place method APTM~\cite{yang2023towards} by $0.68 \%$, $0.64 \%$, and $0.31 \%$. 
Although our mAP is lower than RaSa~\cite{bai2023rasa}, this is because RaSa employs a mixed training strategy of strong and weak positive sample pairs. This strategy gives a larger boost to mAP. For a fair comparison with other methods, we do not use this strategy. As a reference, if we use the mixed training strategy for fine-tuning, the mAP can reach $72.14 \%$.
For ICFG-PEDES, we achieve $68.93 \%$ R@1, obtaining $0.42 \%$ and $3.65 \%$ improvements respectively compared to APTM~\cite{yang2023towards} and RaSa~\cite{bai2023rasa}, while all the other recall rates outperform previous methods.
On RSTPReid, we obtain $68.15 \%$ R@1, and $53.77 \%$ mAP, higher than all the other methods.
At the same time, our method has achieved competitive results, \ie, $86.30 \%$ R@5 and $91.10 \%$ R@10, compared to the previous two best methods.

\begin{table}[!t]
\small
\centering
\caption{Comparison with other methods on the domain generalization task. We adopt CUHK-PEDES (denoted as C) and ICFG-PEDES (denoted as I) as the source domain and the target domain in turn.}
\renewcommand\arraystretch{1.2}
\setlength{\tabcolsep}{5pt}
\begin{tabular}{c|l|ccc}
\hline
                                                   & Method                            & R1        & R5       & R10          \\
\hline
\multirow{6}{*}{\rotatebox{90}{C $\rightarrow$ I}} & Dual Path \cite{zheng2020dual}    & 15.41     & 29.80      & 38.19       \\
                                                   & MIA \cite{niu2020improving}       & 19.35     & 36.78      & 46.42       \\
                                                   & SCAN \cite{lee2018stacked}        & 21.27     & 39.26      & 48.83       \\
                                                   & SSAN \cite{ding2021semantically}  & 29.24     & 49.00      & 58.53       \\
                                                   & LGUR \cite{shao2022learning}      & 34.25     & 52.58      & 60.85       \\
                                                   \cline{2-5}
                                                   & \textbf{MRA (Ours)} & \textbf{50.01} & \textbf{67.13} & \textbf{74.24}  \\
\hline\hline
\multirow{6}{*}{\rotatebox{90}{I $\rightarrow$ C}} & Dual Path \cite{zheng2020dual}    & 7.63      & 17.14      & 23.52       \\
                                                   & MIA \cite{niu2020improving}       & 10.93     & 23.77      & 32.39       \\
                                                   & SCAN \cite{lee2018stacked}        & 13.63     & 28.61      & 37.05       \\
                                                   & SSAN \cite{ding2021semantically}  & 21.07     & 38.94      & 48.54       \\
                                                   & LGUR \cite{shao2022learning}      & 25.44     & 44.48      & 54.39       \\
                                                   \cline{2-5}
                                                   & \textbf{MRA (Ours)} & \textbf{47.66} & \textbf{68.70} & \textbf{76.82}  \\
                                                   
\hline
\end{tabular}
\label{tbl: dg}
\end{table}

\noindent \textbf{Qualitative Results.} 
We further provide qualitative text-based person retrieval results of our method and Baseline on CUHK-PEDES (see Fig.~\ref{fig: result}).
The four retrieval results at the top of the image represent successful cases, while the two retrieval results at the bottom are failure cases. Compared to the baseline, our method can distinguish finer-grained features and retrieve more matched person images. We observe that the mismatched images retrieved in failure cases generally still align broadly with the text query descriptions. This is primarily because real-world datasets are influenced by annotator subjectivity during the labeling process, making it challenging to comprehensively describe all characteristics of person images. 
To address this limitation, our future work will explore two directions: (1) leveraging large-scale Vision-Language Models (VLMs) for automated data augmentation to generate more diverse and fine-grained textual descriptions, and (2) designing noise-robust loss functions that can learn effectively from partially accurate or subjective annotations.

\begin{table}[!t]
\caption{Ablation Study on the SDA pretraining dataset on CUHK-PEDES. 
We report fine-tuned recall rates across different pretraining methods. * We reimplement the method with the official code.
}
\label{tab: sdaabl}
\footnotesize
\centering
\begin{tabular}{l|c|c|cc}
\shline
Method & Pretraining Dataset & \# Data & R@1 & R@10 \\
\hline
\multirow{4}{*}{IVT*~\cite{shu2022see}} 
& -                           & 0M    & 61.99 & 87.69 \\
& COCO~\cite{lin2014microsoft}, \etc. & 4M    & 63.56 & 88.86 \\
& MALS~\cite{yang2023towards} & 1.51M & 64.85 & 89.99 \\
& SDA                         & 1.22M & 65.35 & 89.90 \\
\hline
\multirow{2}{*}{APTM~\cite{yang2023towards}}
& -                           & 0M    & 66.44 & 90.76 \\
& MALS~\cite{yang2023towards} & 1.51M & 76.53 & 94.15 \\
\hline
\multirow{2}{*}{MRA (Ours)}
& -                           & 0M    & 71.23 & 92.43 \\
& SDA                         & 1.22M & \textbf{77.21} & \textbf{94.46} \\
\shline
\end{tabular}
\end{table}

\noindent \textbf{Domain Generalization.}
We conduct a series of domain generalization experiments to verify the generalization ability of MRA, following LGUR~\cite{shao2022learning}. As shown in Table \ref{tbl: dg}, we use models trained on the source domain to evaluate performance on the target domain. The CUHK-PEDES dataset and the ICFG-PEDES dataset serve as the source and target domains in turn. 
From CUHK-PEDES to ICFG-PEDES, MRA achieves SOTA results, \ie, $50.01 \%$ R@1, $67.13 \%$ R@5, and, $74.24 \%$ R@10, against other methods. 
On the task of domain generalization from ICFG-PEDES to CUHK-PEDES, MRA still outperforms all the other methods such as LGUR~\cite{shao2022learning} by a significant margin ($+22.22 \%$ R@1 improvement).

\begin{figure*}[!t]
\centering
\includegraphics[width=0.72\linewidth]{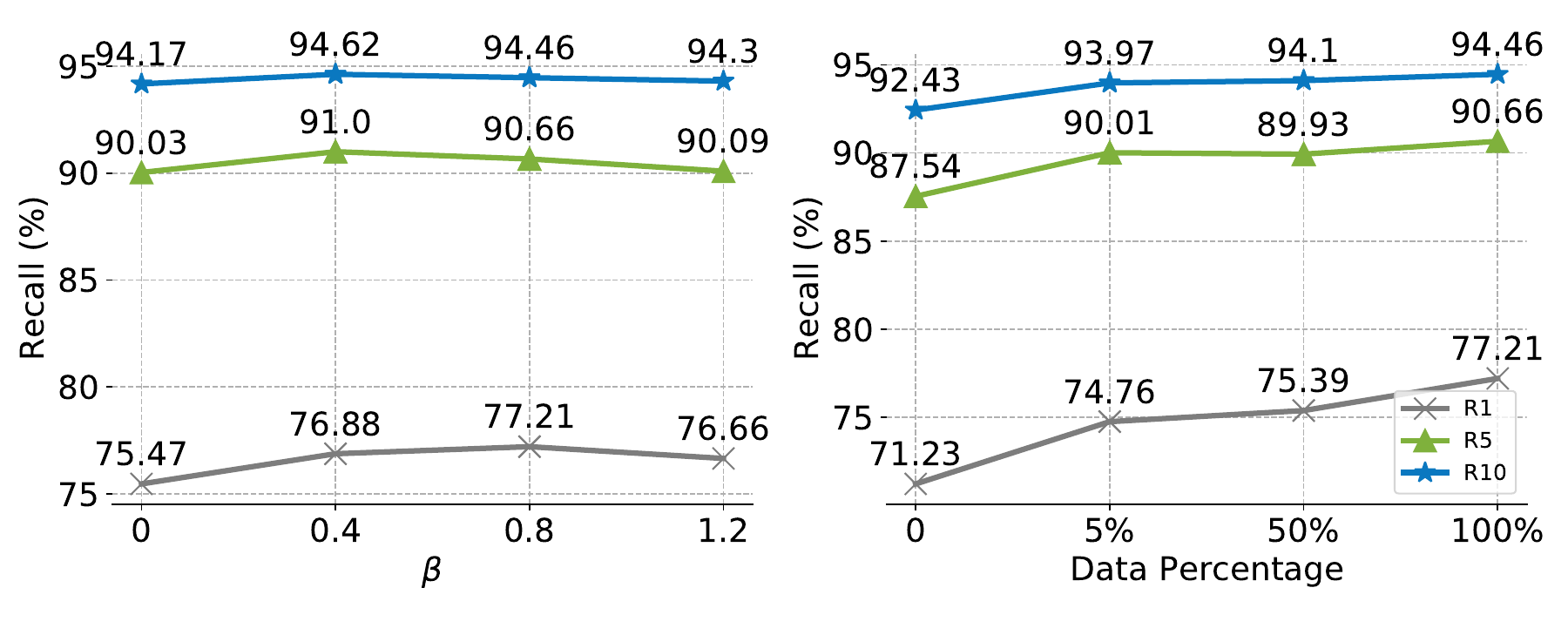}
\caption{Ablation Study on $\beta$ and the pretraining data scale. In left subfigure, we set $\beta = \{0, 0.4, 0.8, 1.2\}$ respectively to validate the impact of $\beta$ on MRA. 
In right subfigure, we apply $0\%$, $5\%$, $50\%$, and $100\%$ data of SDA to pretrain. The fine-tuned recall rates on CUHK-PEDES are reported.  
}
\label{fig: beta_&_scale}
\end{figure*}

\noindent\textbf{Are SDA and MRA beneficial for image-based person re-identification?} 
We further evaluate the scalability of MRA on traditional image-based person re-identification tasks. 
Specifically, we compare the performance of the Swin Transformer (Swin-Base)~\cite{liu2021swin} pretrained on SDA (MRA Vision Encoder) with that pretrained on ImageNet-1K~\cite{deng2009imagenet} on the Market-1501~\cite{zheng2015scalable} dataset. 
As shown in Table~\ref{tab: reid}, under the same fine-tuning hyperparameter settings (image size: $224 \times 224$, fine-tuning epochs: 60), the MRA Vision Encoder achieves $94.39 \%$ R@1 and $82.46 \%$ mAP, outperforming the ImageNet-1K~\cite{deng2009imagenet} pretrained Swin Transformer by $+3.27\%$ R@1 and $+7.95 \%$ mAP. 
The experimental results show that the SDA dataset brings significant improvements to conventional image-based person re-identification, indicating strong robustness of MRA.

\begin{table}[!t]
\caption{Ablation Study on MRA loss on CUHK-PEDES. We apply $68,126$ data pairs from SDA to pretrain and then report the fine-tuned recall rate on CUHK-PEDES. $\mathcal{L}_{v1}$, $\mathcal{L}_{v2}$ and $\mathcal{L}_{v3}$ are losses of three variants of $\mathcal{L}_{MRA}$.
}
\label{tab: abl}
\footnotesize
\centering
\begin{tabular}{l|cccc|cc}
\shline
Method & $\mathcal{L}_{v1}$ & $\mathcal{L}_{v2}$ & $\mathcal{L}_{v3}$ & $\mathcal{L}_{MRA}$ & R@1 & R@10 \\
\hline
M1   & & & &              & 73.99 & 93.57 \\
M2   & $\checkmark$ & & & & 74.04 & 93.58 \\
M3   & & $\checkmark$ & & & 74.19 & 93.54 \\
M4   & & & $\checkmark$ & & 74.43 & 93.82 \\
\hline
MRA (Ours)  & & & & $\checkmark$ & \textbf{74.76} & \textbf{93.97} \\
\shline
\end{tabular}
\end{table}

\section{Ablation Studies}

\noindent \textbf{Effectiveness of Pretraining.} 
To validate the effectiveness of pretraining, we conduct a comprehensive empirical analysis on CUHK-PEDES. The experimental results are displayed in Table~\ref{tab: sdaabl}, where the accuracy of R@1 and R@10 is reported (To save space, R@5 and mAP are not reported).
The results in the last two rows of the table, \ie, the experimental results of Baseline \emph{VS.} MRA, show the impact of pretraining. After pretraining on SDA, MRA achieves $5.98 \%$ R@1 and $2.03 \%$ R@10 improvement. 
Similar results could be observed on ICFG-PEDES and RSTPReid, as shown in Table~\ref{tab:sota_ICFG} and Table~\ref{tab:sota_RSTP}. 
To intuitively show the advantages of pretraining, we present the qualitative comparison results of Baseline \emph{VS.} MRA in Figure~\ref{fig: result}.

\noindent \textbf{The Impact of SDA.} 
To further illustrate the superiority of SDA as a pretraining dataset, we compare it with MALS~\cite{yang2023towards}, a pedestrian image-text dataset with rich attribute annotations (used in APTM~\cite{yang2023towards}), and the general image-text datasets COCO~\cite{lin2014microsoft}, \etc, which contains four image caption datasets (refer to IVT~\cite{shu2022see} for details), totaling $4M$ image-text pairs. 
As shown in Table~\ref{tab: sdaabl}, pretraining on either MALS~\cite{yang2023towards} or SDA significantly improves retrieval accuracy. 
Notably, our MRA, pretrained on less data, outperforms APTM~\cite{yang2023towards} by $0.68\%$ in fine-tuned Recall@1.
Due to the architectural specificity of MRA and APTM~\cite{yang2023towards}, we further compare the pretraining datasets on IVT~\cite{shu2022see}, a general text-based person retrieval method that learns global image-text alignment without leveraging explicit attributes or local regions. 
For a fair comparison, we reproduce IVT~\cite{shu2022see} under our experimental setup. The results reported in Table~\ref{tab: sdaabl} are from our implementation.
The results show that IVT~\cite{shu2022see} pretrained on SDA achieves $65.35\%$ R@1, surpassing models pretrained on MALS~\cite{yang2023towards} and COCO~\cite{lin2014microsoft}, \etc.

\noindent\textbf{Effectiveness of MRA Loss.} 
MRA promotes the accuracy of pedestrian retrieval through multi-granularity relation alignment. 
For image-text level relation alignment, the effectiveness of ITC, ITM, and MLM among them has been demonstrated in previous work~\cite{jiang2023cross, bai2023rasa, yang2023towards} and will not be repeated in this paper. 
We focus on the effect of region-phrase level relation alignment.
In this part, the pretraining data scale is set to $68,126$ (the same as the training set of CUHK-PEDES). This setting is mainly for fair comparison and to avoid time-consuming experiments.
As shown in Table~\ref{tab: abl}, Method M1 only adopts image-text level relation alignment, without region-phrase level relation alignment.
The experimental results of our method and Method M1 indicate the performance improvement of MRA ($+0.77 \%$ R@1 and $+0.40 \%$ R@10). 
In other words, simultaneous relation alignment at multiple granularities during pretraining facilitates downstream pedestrian retrieval.
To further elucidate the effectiveness of our designed MRA, we compare it with three different variants. 
Specifically, we investigate the effects of image horizontal flipping in Grounding DINO, and finer-grained relation alignment, \ie, object-word level alignment. 
As shown in Table~\ref{tab: abl}, both variant 1 ($\mathcal{L}_{v1}$) and variant 3 ($\mathcal{L}_{v3}$) horizontally flip images at the rate of $50 \%$ in detecting objects and regions. 
For object-word level annotations, we use a string of the most frequently occurring words in CUHK-PEDES as the input prompt in Grounding DINO.
Variants 1 ($\mathcal{L}_{v1}$) and 2 ($\mathcal{L}_{v2}$) both conduct object-word level alignments, with the distinction that variant 2 utilizes only the bounding boxes without horizontally flipping in detecting objects. 
As shown in these experiments, leveraging region-phrase level (without horizontally flipping) annotations to conduct region-phrase level relation alignment yields better performance for cross-modal retrieval, with R@1 reaching $74.76\%$.

\noindent\textbf{Parameter Sensitivity of $\beta$.} 
To explore the effectiveness of the value of $\beta$ in Eq. \ref{eq: L_MRA}, we conduct further ablation experiments. 
We set $\beta$ to $0, 0.4, 0.8,  1, 2$ for pretraining separately and fine-tune the model on CUHK-PEDES. 
As shown in the left figure of Fig.~\ref{fig: beta_&_scale}, if $\beta = 0.4$ or $0.8$, the performance of the model is better. Intuitively, we set $\beta = 0.8$ without loss of generality.

\noindent\textbf{The Impact of Pretraining Scale.} 
The size of the pretraining dataset typically affects the performance of the pretrained model. Increasing the size of the data used in pretraining generally leads to varying degrees of improvement on downstream tasks, whether it is a general vision language pretraining model~\cite{radford2021learning, li2021align} or a text-based pedestrian retrieval pretraining model~\cite{yang2023towards}. In this paper, we further investigate the effect of the SDA scale on the pedestrian retrieval task. Specifically, we conduct ablation experiments varying the size of SDA, with pretraining data amounts set to $0 \%$ (no pretraining), $5 \%$, $50 \%$, and $100 \%$ of SDA, respectively. As shown in the right figure of Fig.~\ref{fig: beta_&_scale}, increasing the amount of pretrained data leads to improvements in the recall rate in line with previous research. Notably, there is a significant improvement in fine-tuning performance from $0 \%$ to $5 \%$, while the rate of performance improvement decreases from $5 \%$ to $100 \%$.
Our experiments indicate that a model pretrained on merely $5\%$ of the data achieves approximately $97\%$ of the performance achieved after fine-tuning on the downstream task. We attribute this phenomenon to the high degree of diversity and representativeness inherent in a small, randomly sampled subset of our pretraining corpus. Analysis reveals that this subset comprehensively covers the vast majority of person attributes, background, and linguistic structures present in the full SDA dataset. Furthermore, the pretraining dataset was deliberately constructed through domain adaptation to align closely with the target downstream CUHK-PEDES dataset, resulting in a high degree of domain correlation. Consequently, the feature representations learned from this limited pretraining data are already highly transferable and sufficient for effective adaptation to the target tasks.

\section{Conclusion}
\noindent In this work, we have systematically analyzed the significant discrepancies between synthetic pretraining datasets and real-world pedestrian retrieval datasets, largely attributed to variations in illumination, pose, and other visual characteristics. 
To mitigate these domain shifts, we introduce a unified framework that tackles the domain gap at both the image and region levels. At the image level, we employ Domain-aware Diffusion (DaD) to reduce the stylistic and distributional divergence between the synthetic pretraining data and the real-world target dataset. This approach facilitates the effective utilization of synthetic data for pretraining, thereby enhancing the generalization capabilities of pedestrian retrieval models to real-world conditions.
At the region level, we augment the dataset by integrating region-phrase annotations with image-text pairs, thereby strengthening the semantic correspondence between visual content and descriptive text. We propose a Multi-granularity Relation Alignment (MRA) approach to fully leverage these enriched image-text pairs. MRA ensures comprehensive alignment between pedestrian images and corresponding text across different granularities, from coarse to fine, thus addressing the fine-grained discrepancies often challenging cross-domain retrieval.
The effectiveness and robustness of our dual-level adaptation approach, encompassing DaD and MRA, have been thoroughly validated through extensive experiments on three benchmark pedestrian retrieval datasets, \ie, CUHK-PEDES, ICFG-PEDES, and RSTPReid. 
Despite the promising results, we acknowledge the limitations of our approach. First, although DaD generates a large volume of domain-aligned images, it still inherits the inherent drawbacks of diffusion models, such as oversaturation in some generated images. Second, the captions for these images are produced by the BLIP2 model, a process that may introduce and propagate the inherent biases present in BLIP2.
Looking forward, our future work will explore several promising directions. A primary focus will be to extend this research from images to video-based person retrieval, investigating feasible solutions in more complex and realistic scenarios. This includes studying the impact of various human behaviors, especially abnormal activities~\cite{yang2025beyond}, on retrieval performance. We believe tackling these dynamic and complex settings is a crucial step toward real-world application.

\noindent\textbf{Acknowledgments.}
This work is supported by National Key Research and Development Program of China (2023YFC3321600).
Z. Zheng acknowledges supports from Guangdong Basic and Applied Basic Research Foundation 2025A1515012281, the University of Macau MYRG-GRG2024-00077-FST-UMDF, the University of Macau Advanced Research Institute in Hengqin, and the Macao Science and Technology Development Fund Grant FDCT/0043/2025/RIA1.

{
    \small
    \bibliographystyle{ieeenat_fullname}
    \bibliography{main}
}

\end{document}